\documentclass[10pt,journal]{IEEEtran}
\usepackage{todonotes}

\ifCLASSOPTIONcompsoc
  \usepackage[nocompress]{cite}
\else
  \usepackage{cite}
\fi
\ifCLASSINFOpdf
\else
\fi
\usepackage[normalem]{ulem}
\usepackage{babel}
\usepackage{amsmath}

\usepackage{graphicx}
\usepackage{fancyhdr}
\usepackage{multirow}
\usepackage{booktabs}
\usepackage[normalem]{ulem}

\begin{document}
%
\title{Detecting Nighttime Anomalies from NASA Black Marble Using a Generalized Spatio-Temporally Robust Framework of Machine Leaning Ensembles

 }
%
%
%
%
\author{Srija~Chakraborty\\
\textit{Universities Space Research Association}\\
schakraborty@usra.edu}

\IEEEtitleabstractindextext{
\begin{abstract}
Nighttime lights from NASA's Black Marble product suite capture thermal and light emission signals from anomalous events including fires, volcanic eruptions, and gas flaring. Existing detection approaches rely primarily on thermal bands, limiting sensitivity to weaker signals. We propose a novel machine learning framework that jointly models Black Marble M-band and Day/Night Band (DNB) signals to derive a generalized, spatio-temporally robust ensemble of anomaly detectors. The framework iteratively builds detectors that scale across regions, seasons, anomaly classes, and extends over land and ocean. Detection sets at varying confidence levels are derived based on relevant bands and detector agreement. The approach improves true detection rate while reducing spurious detections and results demonstrate strong generalizability with applications in natural hazard monitoring and energy extraction.
\end{abstract}}
\maketitle

\IEEEdisplaynontitleabstractindextext
\vspace{2em}

%
\IEEEpeerreviewmaketitle

\IEEEraisesectionheading{\section{Introduction}\label{sec:introduction}}

%
%
%
%

\IEEEPARstart{R}{emotely} sensed nighttime lights (NTL) from NASA's Black Marble product~\cite{roman2018nasa} capture a range of phenomena relevant for understanding and tracking the Earth system at night. These range from natural phenomena such as fires~\cite{polivka2016improving}, aurora~\cite{kalb2023aurora} to anthropogenic events such as gas flaring~\cite{chakraborty2023potentially}, urbanization~\cite{xie2019temporal},\cite{chakraborty2023adaptive},\cite{li2026satellite}, disasters impacting power grids~\cite{wang2018monitoring},\cite{roman2019satellite},\cite{zhao2020time},\cite{chakraborty2023adaptive}, electrification~\cite{chakraborty2025global}, socio-political conflicts and displacement~\cite{bennett2017advances},\cite{li2013satellite},\cite{jiang2017ongoing},~\cite{chakraborty2023adaptive}, and impacts of unprecedented events such as COVID-19~\cite{stokes2022tracking},~\cite{chakraborty2025globalCov} serving a diverse range of applications~\cite{levin2020remote}. Black Marble NTL data, derived from the Visible Infrared Imaging Radiometer Suite (VIIRS)~\cite{justice2013land}, offer a unique vantage point for observing daily signals related to these events globally due to the high revisit rate of the SNPP and NOAA-20 satellites. The daily revisit is especially salient for capturing both persistent and transient events while the long-term record from VIIRS allows cataloging these occurrences over time. In particular, the daily top-of-atmosphere Black Marble data at a moderate spatial resolution of 500 m record both natural and anthropogenic anomalous events such as fires, volcanic eruptions, and gas flaring from energy extraction on land and water. VIIRS data have been used for detecting these events based on VIIRS thermal (Moderate M) bands capturing the thermal emission signal. The VIIRS Day/Night Band (DNB) that lies between the visible and Near-Infrared regions has a large dynamic range and captures the light emission signal of the events and are sensitive to weaker instances with less distinct thermal signals~\cite{roman2018nasa},~\cite{chakraborty2023potentially}. Although the DNB has improved flare and fire detection~\cite{polivka2016improving},~\cite{elvidge2013viirs},~\cite{elvidge2019extending}, it is used as a confirmatory feature only. However, the combined use of thermal (M-bands) and light (DNB) emission signals was shown to improve the detection of these events compared to existing methods (VIIRS 375 m Active Fires and VIIRS NightFire) that use a ``thermal band only" approach~\cite{chakraborty2023potentially}. This study developed a machine learning anomaly detector to model the non-anomalous background pixels, creating a threefold advantage: first, it used unsupervised learning to model for background variation that extracted anomalous pixels and accounted for lack of high-quality labels; second it jointly used all Black Marble bands (six M-bands and DNB) to improve detection (higher detection confidence, weaker signal detection, lower false detections); and third it created an independent detection approach, allowing uncertainty estimations when assessed with existing rule-based thermal band approaches. However, the background modeling was localized and region-specific (Eagle Ford Shale, United States) that limits scalability across geographies and seasons.

In this study, we propose a novel suite of spatio-temporally robust, generalized detectors of thermal and light emission anomalies from Black Marble NTL that combines the M-band and DNB signals and derives an ensemble classification of pixels. Our approach builds on the localized method and iteratively constructs the detection workflow leveraging unsupervised detections across regions and seasons. Our modeling builds an effective representation of the large spatio-spectro-temporal variability of the background to detect events. The detectors are generalized, implying geographic and seasonal scalability as well as class-agnostic detections of thermal and light emission anomalies, spanning fires, volcanoes, and gas flares. The approach scales both across the land and ocean. We apply the detectors across known energy extraction sites, retrospectively on fires and volcanic eruptions for demonstration and derive a suite of detections. Accurate ground truth data of these events are limited and thus we use agreement across multiple detectors from our suite to derive a 'high-confidence' data layer. We observe high overlap between each detector and the merged layer. We also analyze the detections to rank the detectors most relevant for operational use. The DNB layer's higher sensitivity allows weaker signal detection missed by M-bands and we note these as a moderate-confidence 'DNB-only' set. 
\begin{figure*}[h]
\centering
\resizebox{5.2in}{2.2in}{\includegraphics{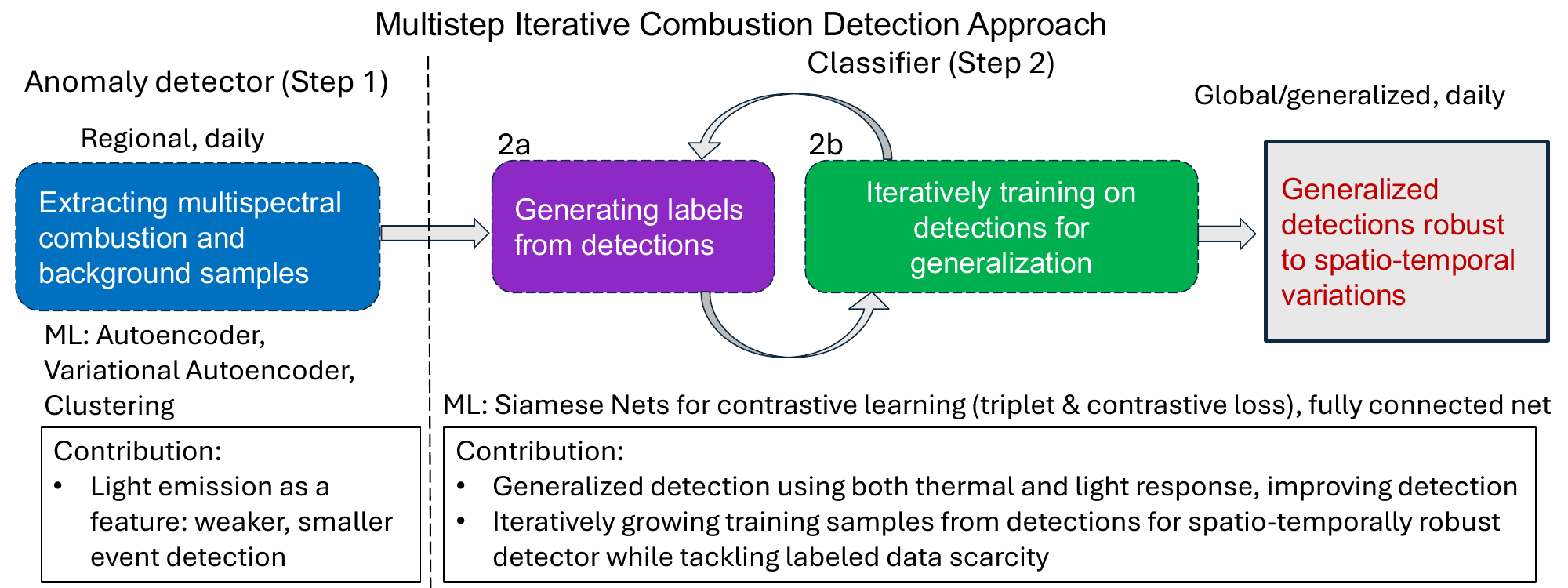}}
\vspace*{-.2mm}
\caption{Proposed two-step iterative anomaly detection approach consisting of anomaly detection, iterative learning with contrastive learning and baseline model ensembles (self-generated labels from detection) for generalization.}
\vspace*{-.5mm}
\label{fig:f1_flowchart}
\end{figure*}
Overall, the contributions of this study are as follows:
\begin{itemize}
    \item a spatio-temporally robust, generalizable thermal and light emission anomaly detection suite using an iterative framework of machine learning ensembles. The approach is globally scalable, extends to land and ocean observations and applicable across classes, including fires, volcanoes, gas flares.
    \item detection sets at varying levels of confidence based on input bands used and agreement between models to form an ensemble.
    \item improves detection performance and weak signal detection by jointly modeling M-bands and DNB while reducing spurious detections.
    \item uses the highest quality NTL, the Black Marble product, making it the first approach to derive Black Marble top-of-atmosphere nighttime anomalies with applications in natural hazard, disaster monitoring, and energy extraction.
\end{itemize}

\section{Methodology}
\subsection{Iterative multi-step approach}

As these events of interest result in a high thermal and light emission signal compared to the background but there is scarcity of high-quality labels, out proposed method takes a multi-step approach. First we use an anomaly detection approach that models the non-anomalous background spectra from a small volume of unlabeled data and monitor deviations from the expected DNB radiance and M-band brightness temperature for detections. This addresses label data scarcity and avoids training on existing thermal band thresholding-based datasets that will bias the detectors to these thresholds and restrict it to a subset of bands only.
The detected anomalies localize possible anomalies and serve as labels required to train a generalized detector that is robust to spatial and temporal variations.
By repeating the anomaly detection step over known event sites, we create a repository of extracted anomalies which then allows refining the detectors iteratively. 
We explore different classifiers that learn to distinguish between the two classes (background and anomaly) and use filtering and masking strategies to remove possible contamination to derive a high confidence label set detected by multiple models. These classifiers initially learn regional models trained locally where the anomaly detectors were applied. We then extended to unseen regions by training jointly on detections from different regions and seasons for generalization. Both our anomaly detectors and iteratively-trained classifiers generate two classes of detections, i) with both thermal and light emission signals and are of high confidence, ii) urban-masked DNB emission-only set that is of moderate confidence.

The proposed methodology has the following steps as shown in Figure~\ref{fig:f1_flowchart}:
\begin{enumerate}
    \item Anomaly detection (step 1): This step creates local background models of the expected thermal and light emission signal in a given region and monitors deviations in subsequent observations to extract pixel-level multi-spectral signals. 
    \item Generalized anomaly detector (step 2): 
    \begin{enumerate}
        \item The detections from step 1 are then utilized as labels to automate annotating the background and anomalous classes. Anomalous signals can be categorized depending on the subset of bands that respond and are related to event class. At this stage we retain labels for binary classification.
        \item Model-annotations are used to iteratively create a generalized detector. We first use the step 1 detections (set-1) as labels to create regional detectors that are applied to a small unseen set of observations from the region (set-2). The models are then refined or retrained on detections of both sets 1 and 2 to create the final generalized regional detector. These are applied to incoming data (test set) over the peak occurrence seasons to examine the performance of the classifiers trained on machine learning generated labels. For both sets 1 and 2 we apply filtering functions to remove contamination to the best of our knowledge. The set 1 and 2 detections of the final regional detectors are jointly trained (across VNP46 tiles) to create a generic model that are then applied to unseen regions and seasons for examining spatio-temporal generalization. 
    \end{enumerate}
\end{enumerate}
\vspace{-1.2mm}
 While we analyze models from all steps in the workflow, the jointly trained general model (referred as 'joint'), is most important to assess for its suitability to be used operationally.

 \subsection{Input Features}

\textbf{Thermal Infrared Emission}\\
M-10 to M16: We use all VNP46A1 thermal bands to model the non-anomalous background. Anomalous thermal emission spectra are detected based on high error from background model in unseen observations. 

M-10 and M11: As gas flaring peaks in the M-10 and M-11 bands and we observe the peak signal in M-11, we apply the outlier based detection $\mathcal{T}_{10}$, using outliers from the $90$-th percentile of the M-11. As these bands do not capture background properties such as land water, clouds, urban areas, these are ideally suited for extracting anomaly without requiring masking.

The detections from M-10 to M-16 and the outlier detections of M-11 together form the anomalous thermal emission set.

\textbf{Light Emission}\\
Day/Night Band Models (Visible to Near-Infrared Emission):
The DNB captures light emission information. When DNB detections are in agreement with thermal bands, it forms the joint high-confidence set. When no co-occurrent thermal bands signals are observed, we define this as the moderate confidence set, which is filtered using the ancillary bands described below to retain likely anomalies.
\begin{figure*}[t]
\centering
\resizebox{5.2in}{4in}{\includegraphics{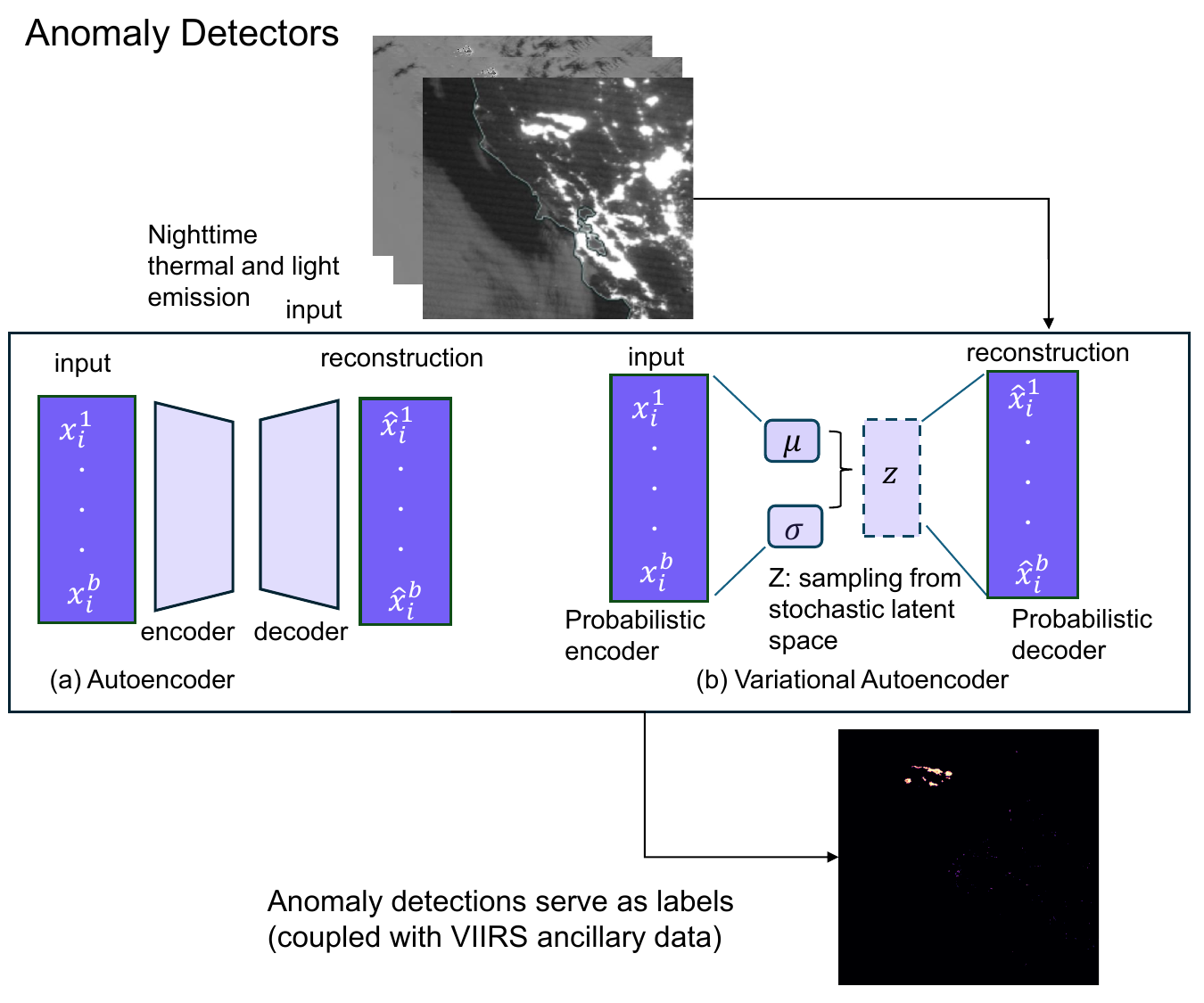}}
\vspace*{-.2mm}
\caption{Proposed step 1 consisting of unsupervised background modeling for anomaly detection to extract labels.}
\vspace*{-.5mm}
\label{fig:f2_step1}
\end{figure*}
\subsection{Machine Leaning models}
At each step in the workflow, we explore relevant architectures to determine suitability. 

\textbf{Anomaly detectors (step 1)}: We use autoencoders (AE)~\cite{hinton2006reducing},~\cite{baldi2012autoencoders} and variational autoencoders (VAE)~\cite{kingma2013auto} to model the background class. We select random pixels from clear-sky nights of regions of interest and train the models to minimize the pixel-level multi-spectral reconstruction error for AE while we used an added KL-divergence term to model the latent space as a standard distribution$~\mathcal{N}(0,\mathcal{I})$. This builds on the approach proposed in\cite{chakraborty2023potentially} and also adds the VAE. We apply to unseen scenes from the same region and detect anomalies based on total reconstruction error. 

These test samples are selected from clear sky-nights but often have slight contamination from clouds in a small fraction of pixels. We observe that magnitude of total reconstruction error is positive and negative for anomalies and clouds, respectively, allowing filtering out contamination. Clouds have lower brightness temperature compared to background, while anomalies deviate positively. To eliminate clouds, we check for the type of per-band deviation where high negative deviations from the background are removed (as possible cloud cover), while high positive deviations are retained as candidate anomalies. This approach was selected over cloud masking as the VIIRS cloud mask (VNP35) is prone to errors at night, which currently mislabels 'nighttime thermal anomalies' as 'clouds'. We also mask detections over water due to its higher nighttime brightness temperature. The remaining anomalies are labeled as `1' and background is labeled as '0'. We repeat this across sites and is shown in Figure~\ref{fig:f2_step1}.

The light emission anomaly from DNB is extracted by partitioning the tile radiance into clusters using a Gaussian mixture model and detecting probable light emission based on a pixel’s cluster assignment and its deviation above the highest mean radiance from the background clusters. We also use an alternative approach where an elastic net models each cluster that defines a central pixel using its neighbors to compute the high radiance likelihood.

\textbf{Generalized spatio-temporal detectors (step 2)}:
As anomaly and background spectra show a large dissimilarity, for our generalized detector (step 2) we explored contrastive learning that learns a representation such that similar pairs are grouped together, while dissimilar pairs are projected far apart. Thus, by training on anomaly and background samples, these methods learn embeddings such that (anomaly, anomaly) and (background, background) pairs are grouped together, while (anomaly, background) pair lies far apart. Our iterative model refining is performed using Siamese networks that contain identical subnetworks that share weights, allowing the comparison of features extracted from both the classes using contrastive and triplet losses ($c$, $t$). The network weights are optimized to extract features from each class and derive embeddings that separate them according to their relative similarity. The set 1 and 2 detections from contrastive learning are also used as labeled data for training a fully connected neural network (FCNN) for comparison. 

A large degree of imbalance exists in the datasets as anomalies are rare. We use all filtered instances of detections (from previous steps) in our anomaly set for training, while we select a random sample of background instances to create the background class. Our network losses are thus weighed inversely based on class frequency to avoid a skewed penalty towards one class.

\textbf{Multispectral Embeddings using Siamese Networks}: Our motivation to use contrastive learning is its ability to utilize pairwise similarities within samples and learn a model that derives embeddings of each class based on its relative distance from each other. As anomaly and background are marked by its dissimilarity, this is ideally-suited for the problem at hand.
To this end, we use a Siamese network that consists of identical subnetworks sharing the same architecture and network weights~\cite{chopra2005learning},~\cite{koch2015siamese},~\cite{zhan2017change},~\cite{wang2019simpleshot}. When a sample pair is passed through it, these identical networks apply the same operation to extract features from the pair, which are then used to train the networks to contrast similar and dissimilar pairs and project the sample to the appropriate feature space based on its class. This generates embeddings of each class and as a result, the network can distinguish between similar and dissimilar pairs, which relates to anomaly and background.

To train the networks, we generate embeddings $\mathcal{E}_{c}$, using contrastive loss and triplet loss $\mathcal{E}_{t}$. The trained networks are then used to generate embeddings of the two classes that maps a sample input (8-D) to the appropriate feature space (2-D). We then use fully connected layers to classify the embedding as anomaly or background. These series of models for generating an embedding and then classifying are represented as $\mathcal{E}_{c}$ and $\mathcal{E}_{t}$ based on the loss. Detections made by these methods on sets 1 and 2 form the final training dataset.

\textbf{Baseline Models}: Additionally, we train an FCNN on the filtered samples of sets 1 and 2 (final training dataset) to classify the samples using a weighted loss. This approach serves as a basleine. 

We also use the most responsive VIIRS thermal band and light emission bands to derive simple outlier (top $10$ and $2$ percentile respectively) detections denoted as $\mathcal{T}_{10}$ (M-11) and $\mathcal{T}_{10}\mathcal{L}_{2}$ (M-11 and DNB), respectively. 

\begin{figure*}[t]
\centering
\resizebox{5.5in}{3in}{\includegraphics{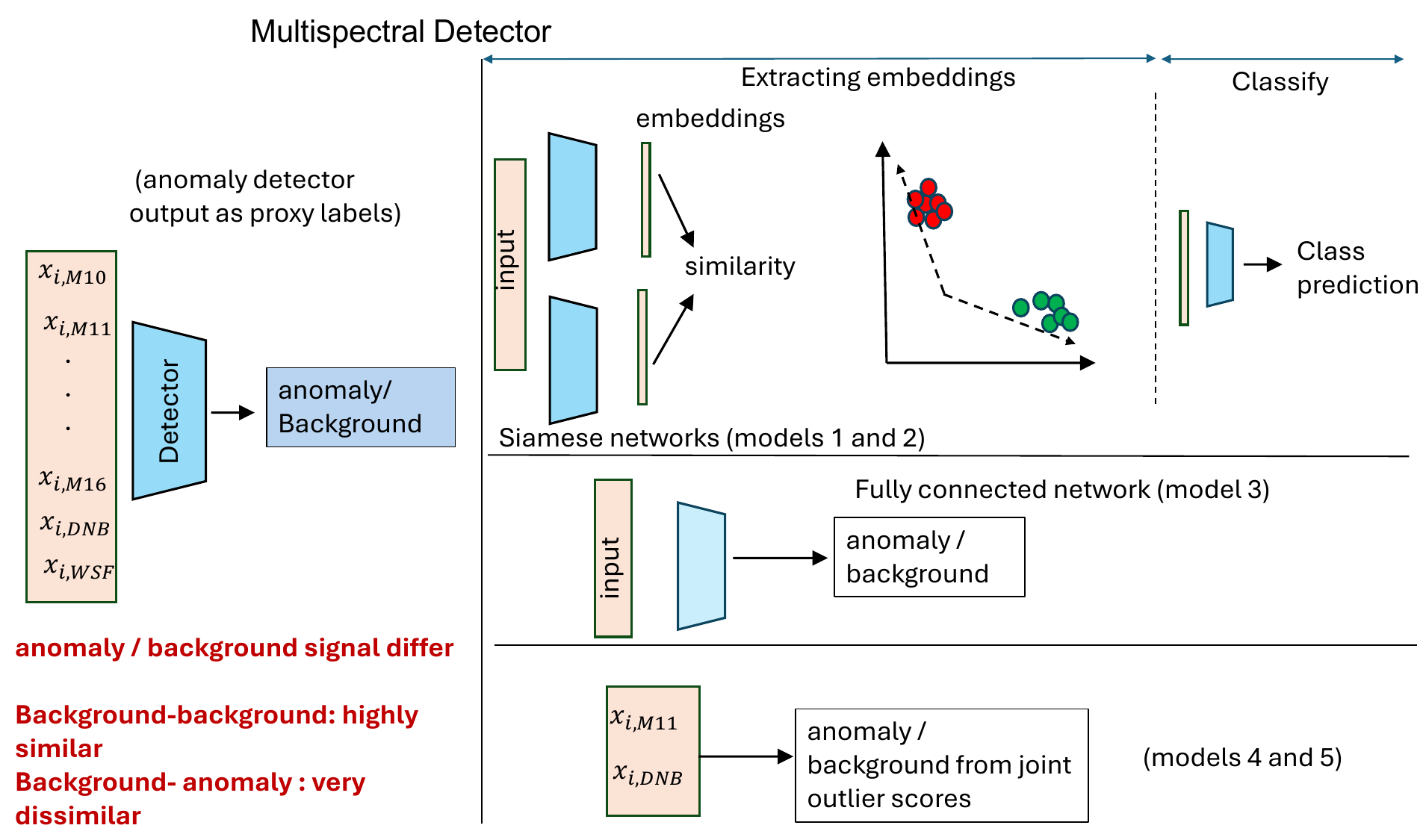}}
\vspace*{-.2mm}
\caption{Proposed step 2 to distinguish anomaly and background using a suite of machine learning and baseline ensembles by leveraging spectral dissimilarity.}
\vspace*{-.5mm}
\label{fig:f3_step2}
\end{figure*}

These creates a total of five detectors as shown in Figure~\ref{fig:f3_step2}. For all detectors, the training set was split to create a validation set where we tuned detection thresholds. In the iterative process, we assessed these thresholds extended effectively to the unseen test set samples at each stage before proceeding to the next step. For each model we set the thresholds to a constant throughout the workflow. 

\textbf{Ancillary Input Layers}:
The AE and VAE from step 1 and $\mathcal{E}_{c}$ $\mathcal{E}_{c}$, FCNN, $\mathcal{T}_{10}$, and $\mathcal{T}_{10}\mathcal{L}_{2}$ from step 2 also capture several background features such as clouds, water, and nightlights from urban areas and settlements. Ancillary VIIRS land-water mask, and World Settlement Footprint (WSF) 2019~\cite{marconcini2020outlining} layers are used to filter out these signals.
The trained models are applied to unseen clear-sky observations to detect anomalous thermal and light emission pixels. We use a small volume of data to train our models and use a validation subset for determining the threshold for the anomaly detectors.

\textbf{Models and Inputs}: The AE and VAE in step 1 uses M-10 to M-16 for the thermal detection set. The DNB detections in step 1 are used to form the light-only detection set. 

In step 2, the Siamese networks $\mathcal{E}_{c}$ $\mathcal{E}_{c}$ and the FCNN require M-10 to M-16, DNB, and the WSF as inputs. $\mathcal{T}_{10}$ requires M-11 as input. $\mathcal{T}_{10}\mathcal{L}_{2}$ requires M-11 and the DNB as inputs.

\begin{figure*}[h]
\centering
\resizebox{5.8in}{3.8in}{\includegraphics{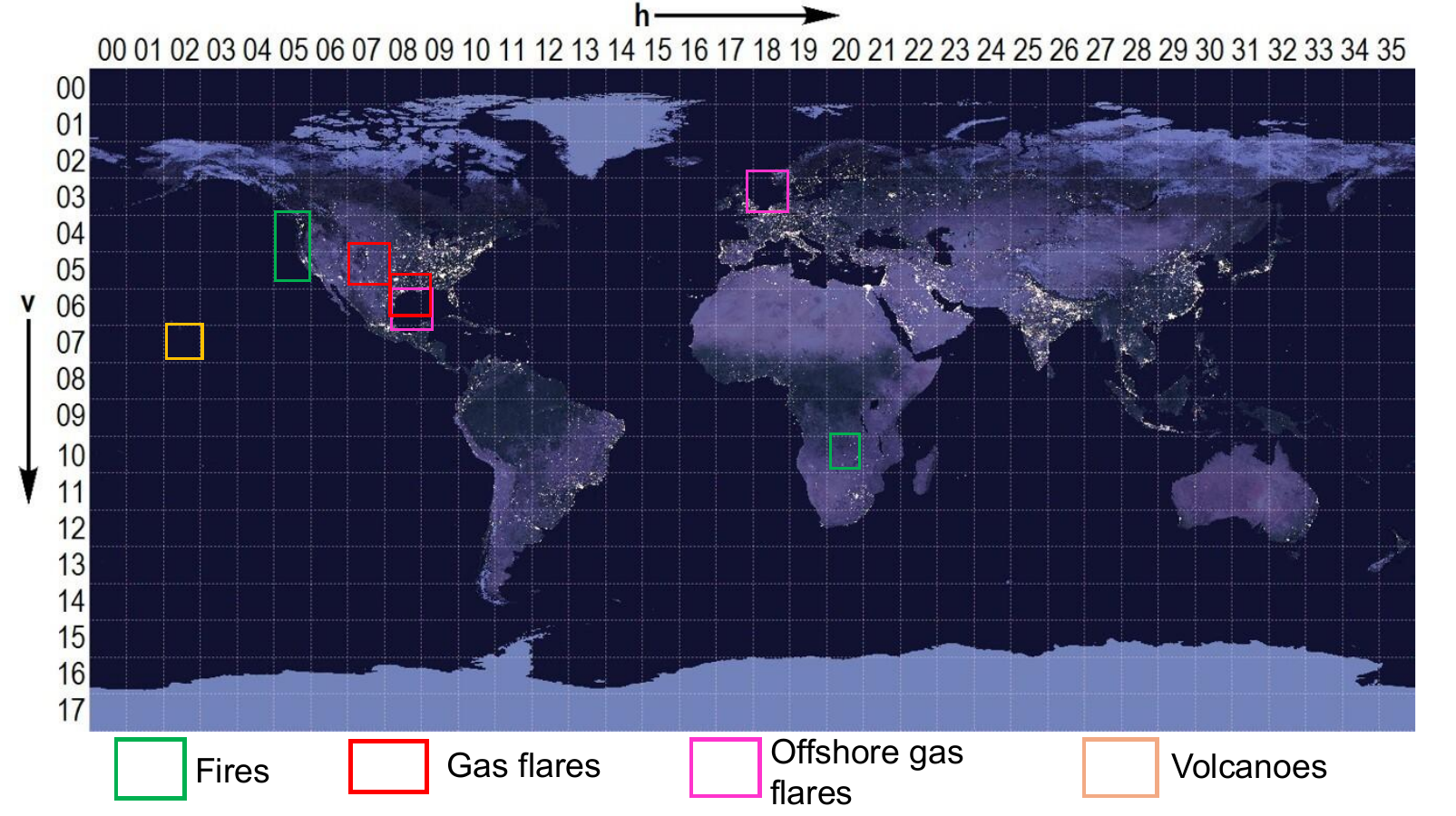}}
\vspace*{-.2mm}
\caption{Proposed regions of study and corresponding Black Marble tiles. Each tile is centered around a region of study where anomaly is prevalent.}
\vspace*{-.5mm}
\label{fig:f4_tiles}
\end{figure*}

\section{Dataset and Regions of Study}
We use the daily, top-of-atmosphere Black Marble observations (VNP46A1) consisting of the Moderate band brightness temperature (six bands; M-10 to M-16) and the visible to Near-Infrared radiance observation from the DNB along with the ancillary cloud and land-water mask, World Settlement Footprint (WSF) 2019~\cite{marconcini2020outlining} and create pixel-based detectors.

Regions of study: The study area consists of the regions (corresponding VIIRS tiles) as shown in Figure~\ref{fig:f4_tiles} that are susceptible to fires, consist of persistent gas flaring sites, and retrospective volcanic eruptions sites. The study duration consists of cloud-free observations from the peak fire season (forest fire: h05v05; 38 nights and controlled burning: h20v10; 92 nights), and brightness temperature extremes (summer and winter) for gas flaring (h08v06; 32 nights in summer and 42 nights in winter). For generalization, we also explored two additional regions for fires in Western United States (h05v04; 53 nights), flaring in the highly emitting Permian basin (h07v05; 77 nights), Kīlauea volcano in Hawaii (h02v07; 231 nights). The anomaly detection and iterative training are performed during the study period over 2018, while the generalization tests are performed in 2019 - 2021. This is shown in Figure~\ref{fig:f4_tiles}.

Our datasets have been fetched from the LAADS repository and our analyses have been performed in a containerized environment in the Nautilus HyperCluster using TensorFlow/Keras.

\section{Metrics}
The step 1 detectors are evaluated for pixel-level co-location with the VIIRS NightFire product similar to~\cite{chakraborty2023potentially}. For step 2, we derive the labels `1' as anomalous and '0' as background. We use a voting approach combing the five detectors to derive the high confidence set. For each detector, if a pixel's predicted label matches the labels of this high confidence set, it counts towards an overlap score. Similarly, for each detector, if a pixel is labeled `1' by a detector but is missed in the high confidence set, we account that in the detector's probability of extra detections.


\begin{figure*}[h]
\centering
\resizebox{4.5in}{4.5in}{\includegraphics{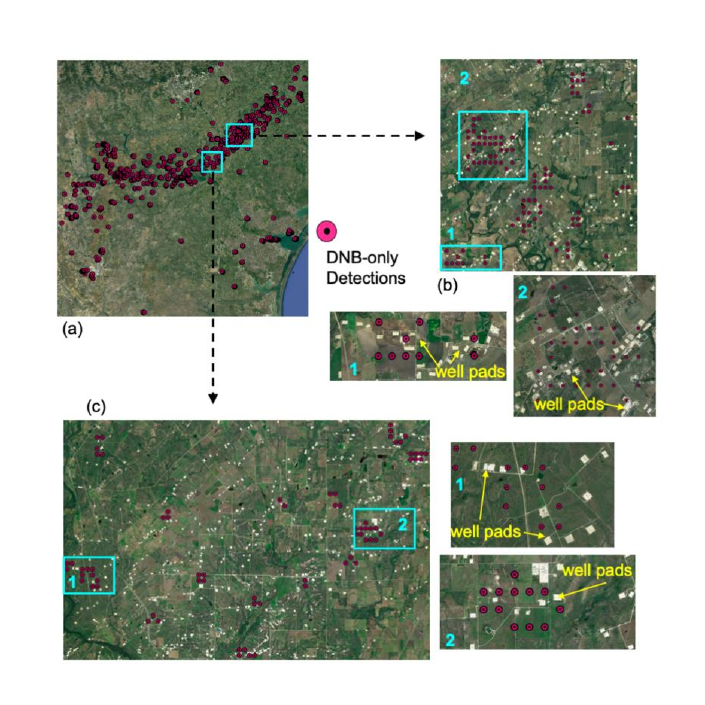}}
\vspace*{-.2mm}
\caption{Anomalous light emission detection from step 1 (DNB-only set) in h08v06 against a regional Landsat composite and (b) and (c) examples of overlap with visible flaring signature (well pads). The high colocation with flare-pads indicates that this set consists of probable flare signals missed by thermal bands and are minimally contaminated from non-flaring sources.}
\vspace*{-.5mm}
\label{fig:f4_flare_pads}
\end{figure*}


\section{Detection Performance and Discussions}

\textbf{Anomaly Detection:}

We compare our detections with the commonly used VNF dataset and compute the overlap between daily VNF detections and from the anomaly detectors, as shown in Table~\ref{ref:table_ad_ov}. 
We observe comparable overlap between autoencoder and variational autoencoder across all regions. The high confidence combined detections show a high overlap with VNF. On visually examining detections in h08v06 (a gas flaring site), we observed that most non-overlaps are across non-confirmed VNF detections. Thus, jointly using all VNP46 retains highly likely anomaly signals and removes possible spurious detections based on single-channel detections. DNB-only light emission signal may be effective in extracting additional detections based on the type of anomaly and its relevance varies across regions based on anomaly source area and spectra peak. For example, we observed high information gain for flaring in h08v06 in the DNB with respect to the thermal bands, while for the flares over h22v05, it is less informative. The joint detections show a higher overlap combining both sets. This results in hierarchies of detection confidence and creating sets of high confidence combined detections and moderate confidence DNB-only detections. Including the DNB-only light emission signal lowers the detection threshold enabling the detection of weaker anomalies that do not have a distinct thermal emission signal without increasing false positives. 

\begin{table}[]
\caption{Overlap of Detected Anomalies with VIIRS Nightfire}\label{ref:table_ad_ov}
\centering
\begin{tabular}{|l|lll|}
\hline
\multirow{2}{*}{Detectors} & \multicolumn{3}{l|}{Overlap o \%} \\ \cline{2-4}
                  & \multicolumn{1}{l|}{h05v05} & \multicolumn{1}{l|}{h08v06} & h20v10 \\ \hline
                 Combined-AE & \multicolumn{1}{l|}{92.66} & \multicolumn{1}{l|}{85.75} & 76.83 \\ \hline
                 Combined-VAE & \multicolumn{1}{l|}{92.87} & \multicolumn{1}{l|}{85.69} & 76.83 \\ \hline
                 DNB & \multicolumn{1}{l|}{77.21} & \multicolumn{1}{l|}{83.46} & 67.58 \\ \hline
                 Joint-AE & \multicolumn{1}{l|}{93.84} & \multicolumn{1}{l|}{96.15} & 84.83 \\ \hline
                 Joint-VAE & \multicolumn{1}{l|}{93.84} & \multicolumn{1}{l|}{96.21} & 84.83 \\ \hline
\end{tabular}
\end{table}


DNB-only signal interpretation: As the DNB is sensitive to weaker anomalies, it increases the fraction of anomalies (flaring: h08v06) observed as shown in Figure~\ref{fig:f4_flare_pads}. However, as these detections lack thermal band signals, we examined a regional Landsat composite to understand the underlying land use and observed a high spatial correlation between these detections and visible signatures of gas flaring infrastructure as shown in Figures 6 b and c. This indicates that due to masking out of urban settlement signals with WSF, these detections are possible flaring activities that are missed by the thermal bands and show a more accurate representation of anomaly in the region. This can potentially increase the detection of weaker fires (small area, early detection) and flares. However, when electric lighting co-occurs with such detections, VIIRS-based detections may not be sufficient to separate contamination from true anomaly.

\textbf{Generalized Model Detections and Detection Confidence}
\begin{table*}[h]
\centering
\caption{Generalized or joint model detection performance across tiles recording overlap (o) and probability of extra detections (p) across all five models. These use all seven Black Marble bands. Bottom three rows show performance on unseen regions. Blue indicates highest performance, magenta shows second highest.}
\label{ref:table_jt_o_P}
\begin{tabular}{|l | ccccc | ccccc|}
\hline
\multirow{2}{*}{Tile}
  & \multicolumn{5}{c|}{overlap $o$}
  & \multicolumn{5}{c|}{probability of extra detections $p$} \\
\cline{2-6} \cline{7-11}
  & o-cont & o-trip & o-fcnn & o-top10 & o-t10\_l
  & p-cont & p-trip & p-fcnn & p-top10 & p-t10\_l \\
\hline\\[-8pt]
h05v05    & \textcolor{blue}{0.9998} & 0.7097 & 0.7252 & \textcolor{magenta}{0.9922} & 0.6495 & \textcolor{magenta}{0.3686} & 0.2092 & 0.1060 & \textcolor{blue}{0.4663} & 0.0311 \\
          & $\pm$0.0001 & $\pm$0.0427 & $\pm$0.0366 & $\pm$0.0026 & $\pm$0.0317 & $\pm$0.0278 & $\pm$0.0452 & $\pm$0.0285 & $\pm$0.0278 & $\pm$0.0128 \\[4pt]
h08v06    & \textcolor{blue}{1.0000} & 0.3247 & 0.9257 & \textcolor{magenta}{0.9621} & 0.9466 & \textcolor{blue}{0.6381} & 0.0000 & \textcolor{magenta}{0.5231} & 0.3485 & 0.0057 \\
          & $\pm$0.0000 & $\pm$0.0125 & $\pm$0.0058 & $\pm$0.0040 & $\pm$0.0052 & $\pm$0.0082 & $\pm$0.0000 & $\pm$0.0089 & $\pm$0.0120 & $\pm$0.0019 \\[4pt]
h08v06\_w & \textcolor{blue}{1.0000} & 0.3724 & 0.9101 & \textcolor{magenta}{0.9650} & 0.9196 & \textcolor{blue}{0.4924} & 0.0270 & 0.3570 & \textcolor{magenta}{0.4192} & 0.0322 \\
          & $\pm$0.0000 & $\pm$0.0198 & $\pm$0.0053 & $\pm$0.0056 & $\pm$0.0081 & $\pm$0.0208 & $\pm$0.0083 & $\pm$0.0219 & $\pm$0.0119 & $\pm$0.0050 \\[4pt]
h20v10    & \textcolor{blue}{0.9999} & 0.5525 & \textcolor{magenta}{0.9992} & 0.9836 & 0.8397 & \textcolor{magenta}{0.3334} & 0.0298 & \textcolor{blue}{0.4820} & 0.2706 & 0.0201 \\
          & $\pm$0.0000 & $\pm$0.0099 & $\pm$0.0002 & $\pm$0.0022 & $\pm$0.0089 & $\pm$0.0122 & $\pm$0.0037 & $\pm$0.0105 & $\pm$0.0105 & $\pm$0.0034 \\[4pt]
h02v07    & \textcolor{blue}{0.9513} & 0.6528 & 0.7183 & \textcolor{magenta}{0.9259} & 0.8009 & 0.1087 & \textcolor{magenta}{0.2202} & 0.0272 & \textcolor{blue}{0.2483} & 0.0880 \\
          & $\pm$0.0134 & $\pm$0.0183 & $\pm$0.0158 & $\pm$0.0148 & $\pm$0.0191 & $\pm$0.0087 & $\pm$0.0173 & $\pm$0.0043 & $\pm$0.0151 & $\pm$0.0095 \\[4pt]
h07v05    & \textcolor{blue}{1.0000} & 0.3846 & 0.9256 & \textcolor{magenta}{0.9736} & 0.9494 & \textcolor{blue}{0.4515} & 0.0692 & 0.3214 & \textcolor{magenta}{0.3285} & 0.0107 \\
          & $\pm$0.0000 & $\pm$0.0100 & $\pm$0.0030 & $\pm$0.0021 & $\pm$0.0036 & $\pm$0.0121 & $\pm$0.0146 & $\pm$0.0113 & $\pm$0.0078 & $\pm$0.0015 \\[4pt]
h05v04    & \textcolor{blue}{0.9995} & 0.8665 & 0.6823 & \textcolor{magenta}{0.9609} & 0.6205 & 0.1168 & \textcolor{blue}{0.3015} & 0.0136 & \textcolor{magenta}{0.2311} & 0.0369 \\
          & $\pm$0.0002 & $\pm$0.0203 & $\pm$0.0134 & $\pm$0.0068 & $\pm$0.0215 & $\pm$0.0119 & $\pm$0.0373 & $\pm$0.0033 & $\pm$0.0212 & $\pm$0.0089 \\
\hline
\end{tabular}
\end{table*}
From the detections by the generalized model we derive a high confidence detection set from our thermal and light bands by utilizing the detections that are found by at least two approaches ($\mathcal{E}_c$, $\mathcal{E}_t$, FCNN, $\mathcal{T_{10}}$, $\mathcal{T}_{10}\mathcal{L}_2$). We show the overlap of each method with this high confidence set in Table~\ref{ref:table_jt_o_P}. The first three tiles show the performance of the region-based training, while h07v05, h05v04, and h02v07 show the performance of the generalization to unseen regions for gas flare, fire detection, and volcanic eruptions respectively. We observe better detection performance by the machine learning methods ($\mathcal{E}_{c}$, $\mathcal{E}_{t}$, FCNN) than the single band approaches ($\mathcal{T}_{10}$ and $\mathcal{T}_{10}\mathcal{L}_{2}$). The performance of the models does not suffer by extending the models across years and seasons (h05v05, h08v06 and h20v10) and to unseen regions (h07v05, h05v04, h02v07) showing high generalizability and robustness to spatio-temporal variations. This comparison is made for the joint, generalized model. 

We observe that $\mathcal{E}_{c}$ has highest overlap score across all regions and event classes, while probability of extra detections is varied, but typically led by $\mathcal{E}_{c}$. This model also has the highest confidence as it uses all bands. This is followed by $\mathcal{T_{10}}$ as it only requires a single band and thus has lower confidence. This is followed by the machine learning baseline FCNN, but it has a higher confidence than $\mathcal{T_{10}}$ and uses all bands. The $\mathcal{E}_{t}$ performs comparably as FCNN for fires and volcanoes but performs very poorly on gas flares and controlled burning (both these are single pixel events covering a very small spatial area). This can be explained by fires and volcanoes have a very high signal in M-10 to M-16 while gas flare signals are constrained to M-10 and M-11 predominantly. The triplet model misclassifies gas flares as it lacks strong signals from M-12 to M-16 and shows the model places strong emphasis on M-12 to M-16 signals as well. A further split of the anomalous classes based on the peak signal bands may improve triplet model performance. 

Our framework also demonstrates the effectiveness of the machine learning-based data labeling approach to alleviate labeled data scarcity that plagues remote sensing and can be extended to extract and model other signals of interest. This methodology demonstrates that by carefully extracting features, small volumes of data can successfully serve as training data and utilizing machine learning-generated labels can address labeled data scarcity in remote sensing.

\subsection{Limitations and Future Extensions}
The primary limitation of monitoring nighttime anomaly is its lack of ground truth data that limits model tuning and evaluation. The proposed approach addresses this by creating multiple models and estimating detection confidence indicated by model agreement. Although this does not serve as ground truth, it serves as an indicator of detector reliability. 
Although anomaly detection is considered as a binary classification problem, multiclass classification could also be explored based on anomaly type indicated through per-band deviation.

Additionally, other model architectures should be explored to comprehensively assess the detectors. 
These include segmentation, augmenting the models with convolutional layers to include spatial information, utilizing a few-shot labeling strategy with the contrastive learning embeddings. Moreover, expert-labeling of the detections may also improve the models with a semi-supervised approach. As the iterative learning process is repeated, the models could also be improved by retraining on missed detection for improving performance. A failed explored approach is a nearest-neighbor classification~\cite{johnson2019billion} of the embeddings due to its computational cost.



\section{Conclusion}
This study proposed a novel spatio-temporally robust framework of machine learning-based nighttime anomaly detection ensembles using NASA's Black Marble thermal and light emission signals to improve true detection rate with a higher confidence. The detectors generalize across regions and seasons and to unseen class types. The approach also extends over land and ocean. We observe that jointly modeling thermal and light emission signals improves detections compared to single band detectors. 
Operationalizing the detectors on daily VIIRS imagery will enable longer-term assessment of model performance and derive a value-added data product for diverse applications such as natural hazard monitoring and energy extraction. 

\section*{Acknowledgments}
Funding for this project was provided by NASA SMD AI/ML Working Group Benchmarking and Training Data Generation Grant and NASA's Multidomain Reusable Artificial Intelligence Tools Grant (80NSSC23K1095). Computa\-tional resources on the National Research Platform were obtained from the Research Institute for Advanced Computer Science, USRA.

\bibliography{refs}
\bibliographystyle{IEEEtran}

\end{document}